# Airplane Detection Based on Mask Region Convolution Neural Network

W.T. Alshaibani[1], Mustafa Helvaci[1], Ibraheem Shayea[2], Hafizal Mohamad[2]

[1]Institute of Informatics, Satellite Communication and Remote Sensing Program, Istanbul Technical University, ITU Ayazaga Campus, Institute of Informatics Building, Sariyer, 34469 Istanbul, Turkey
[2]Electronics and Communication Engineering Department, Faculty of Electrical and Electronics Engineering, Istanbul Technical University (ITU), 34469 Istanbul, Turkey
[2]Faculty of Engineering and Built Environment, Universiti Sains Islam Malaysia, 71800 Bandar Baru Nilai, Negeri Sembilan, Malaysia

Corresponding author: W.T. Al-Shaibani (e-mail: al-shaibani18@itu.edu.tr)

This research has been produced benefiting from the 2232 International Fellowship for Outstanding Researchers Program of TÜBİTAK (Project No: 118C276) conducted at Istanbul Technical University (İTÜ), and it was also supported in part by Universiti Sains Islam Malaysia (USIM), Malaysia.

**ABSTRACT** Addressing airport traffic jams is one of the most crucial and challenging tasks in the remote sensing field, especially for the busiest airports. Several solutions have been employed to address this problem depending on the airplane detection process. The most effective solutions are through the use of satellite images with deep learning techniques. Such solutions, however, are significantly costly and require satellites and modern complicated technology which may not be available in most countries worldwide. This paper provides a universal, low cost and fast solution for airplane detection in airports. This paper recommends the use of drones instead of satellites to feed the system with drone images using a proposed deep learning model. Drone images are employed as the dataset to train and evaluate a mask region convolution neural network (RCNN) model. The Mask RCNN model applies faster RCNN as its base configuration with critical modifications on its head neural network constructions. The model detects whether or not an airplane is present and includes mask estimations to approximate surface area and length, which will help future works identify the airplane type. This solution can be easily implemented globally as it is a low-cost and fast solution for airplane detection at airports. The evaluation process reveals promising results according to Microsoft Common Objects in Context (COCO) metrics.

## I. INTRODUCTION

Traveling is a critical need for humans. People must travel for various reasons such as to improve their lives, escape from nearby dangers, visit relatives, conduct urgent post/parcel transportation or for emergency cases such as help and rescue. Traveling and urgent post/parcel transportation are usually accomplished via airplanes, cars, buses or ships. Nowadays, people mostly prefer to travel by plane. The number of flights performed globally by the airline industry has been steadily increasing since the early 2000s and was expected to reach 40.3 million in 2020 before the covid-19 pandemic. The impact of the covid-19 pandemic has reduced the number of flights to 16.4 million. However in 2021, it is expected to rise to 22.2 million[1]. The dramatic increase in flights will generate plenty of new routes. This will subsequently place tremendous pressure on airports since traffic jams will occur during flights. More flights per day signifies considerable strain on air traffic control. Therefore, a new approach to support air traffic control would be highly advantageous.

Object detection is a process that differentiates between objects based on data extracted from images or videos. Different objects have unique patterns that make them distinct from others. Frames that contain random patterns are considered as noise, thus, the ability to separate these patterns and accurately identify objects is the ultimate goal for target detection. Specialists and image processing experts always strive to identify whether or not an object exists. For example, petrol engineers search for oil in the same areas on earth, epidemiologists explore the causes of disease and meteorologists determine or predict future weather situations based on certain actions or conditions. The basic idea is to compare data with a threshold so as to determine an object's existence or the action that will occur if the limit is surpassed. A simple example would be the detection process in the human visual system. This process searches for signatures that distinguish objects from the available background. The difference between the background and objects must be significant than the threshold in order to be detectable. A clear technological example would be Facebook's automatic tagging; it can directly recognise and identify faces from images. Extensively assessing image processing and understanding its requirements is not only crucial for the classification of different images, but also for determining the precise estimation of object locations and features. This complicated task is the main function of object detection. This function is usually divided into different subtasks: face detection, pedestrian detection and skeletal detection. Object detection provides accurate information for the semantic understanding of images, videos or any type of data.

Due to the rise in user demands, several facilities have been increasing their adoption and dependency on latest technologies. These facilities include the airline industry since most people prefer to fly for several reasons, such as safety and speed. The number of passengers using flight services has dramatically increased, leading to an upsurge in airplanes used. This has resulted in flight traffic congestion. The ground control operations of flights will experience difficulties once traffic increases, therefore, providing a novel method for managing traffic to support the current traffic control is necessary. One of the current methods employed to solve this issue is airplane detection using satellite images with deep learning approaches. The datasets are collected from public satellite sources.

Drone images are normal images taken by cameras attached to drone bodies. Usually, these images can be easily used without any extra processing. These images can be directly forwarded to any system that performs image processing for target detection or switch to detection applications. A well-known system that aids image processing is the deep learning technique. Since drone images are simple images, the enhancement will be great in terms of processing speed. Various features can be implemented within drone cameras to improve their efficiency, such as only capturing moving objects or taking pictures of infrared bodies with high resolution. In contrast, satellite images require complicated pre-processing steps due to the multilayer-structure of each image. Satellite images contain different layers that must be stacked together to form the complete image. Other pre-processing steps are also present to remove noise and make geometric and geographic corrections. Using drone images will reduce the time consumed in performing airplane detection tasks. However, it is illegal to fly drones over airports due to security reasons. Any drone that flies over an airport will be attacked. To utilise drone images, the drones must be placed in specific coordinated points and only used by the airport itself. The employed drones will not fly on a horizontal plane over the airport ground area but on a vertical axis, moving up and down for a specified duration defined by the airport. The main goal is to acquire aerial images in airports. This can even be accomplished by placing 3D cameras on top of several fixed towers. With the help of 3D modelling, the cameras can be used to feed the system with aerial images.

Airport traffic jams can be a nightmare, especially for the busiest airports in the world. Airports usually have a schedule for their daily flights. In the long term, schedules will become overcrowded with flights, and airports will not be able to sufficiently control and manage the issue. Small airports are also expected to have busy schedules in future due to discounts in flight tickets, as noticed in previous years. Some countries are already expanding their airports to meet the vast amount of flight requests. Addressing this problem will help reduce the pressure on airport officers, workers as well as customers. A useful solution with the help of new technology may contribute to rescuing lives or preventing emergency cases. The flight congestion problem is currently being addressed with the utilisation of satellites. Satellites provide images to systems using deep learning techniques to detect airplanes in the airport. However, this method consumes significant processing time and is not an adequate universal solution. Employing satellites to perform one function, such as monitoring an airport, is considered an inefficient economic solution since satellites are extremely costly and require complicated technology. This article proposes the use of drone images with the help of deep learning techniques to build a reliable system that can handle air traffic jams at airports. The cost is much cheaper than satellite-based systems. Drone imaging will allow any country in the world, especially countries that do not possess modern technology or own satellites, to handle flight traffic jams at their airports.

The proposed technique will not just detect whether or not an airplane is present, but it will also determine the surface area of each plane so that traffic control will have accurate information. This data will be integrated in the database of airplane companies to determine the aircraft type according to its surface area. This technique can be applied at any airport around the world. It just requires the presence of drones at specified locations within airports to supply the system with fixed images at certain time periods determined by airport requirements. Compared to drones, satellite techniques are significantly different in terms of cost and complexity. Drones are everywhere, and anyone can obtain them. Satellites, however, are hard to purchase and operate due to complicated requirements. In future, the drone technique can provide the key advantage of precisely determining the surface area of each plane at the airport. This advantage will lead to highly accurate flight control. Drone images do not require extra pre-processing as they are standard images. These images can be directly forwarded to the image detection system. In contrast, satellite images require pre-processing steps for removing noise and conducting graphical modifications and geographical corrections. The drone solution requires less time to perform airplane detection tasks compared to satellite images.

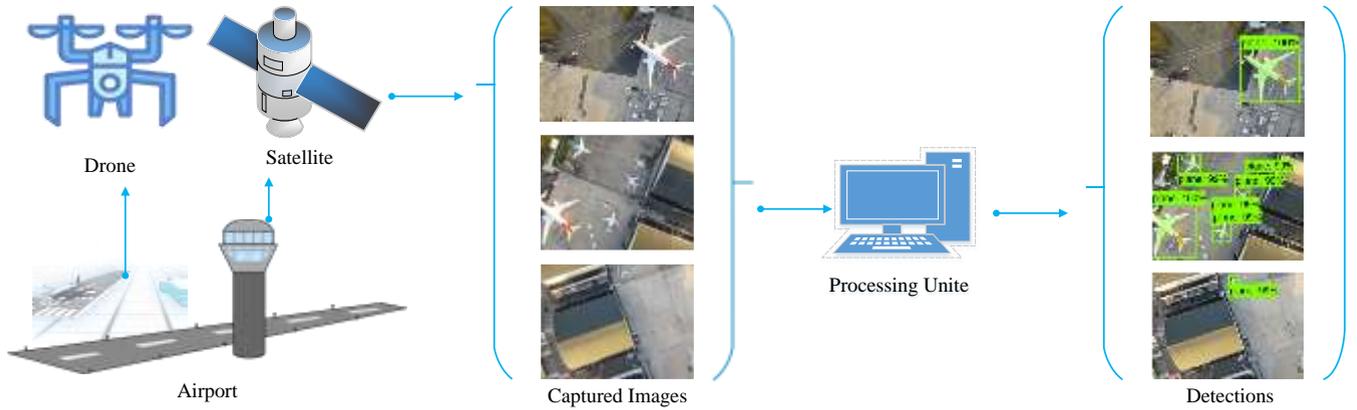
**Figure 1:** Airplane detection process

Weather conditions (such as rain, clouds or snow) will not have the same impact on drone image acquisition and processing stages as satellite-based methods, adding another reliability feature to this approach. By knowing the surface area and length of the airplane, airport officers can determine the airplane type, name, manufacturer and other specifications. In the experimental setup, eight deep learning models have been applied. The Mask RCNN model with the inception backbone model is used as the base model, while the remaining models are modified versions of it, as discussed in the following sections. The variety in models is necessary to determine the optimum hyperparameter values.

The rest of this paper is organised as follows. Section II provides a literature review of related works. Section III explains the methodology and the entire practical steps employed. Section IV discusses the results of the evaluation process. Finally, Section V concludes the paper, subsequently ending with acknowledgments and references.

## II. BACKGROUND AND RELATED WORKS

Remote sensing is the science of collecting an object's data without any physical contact. Observing earth using satellites, planes, ground antennas and drones provides a beneficial ability to collect a wide range of data at different times and during various weather conditions. New technologies that offer high resolution have led to the automatic object detection. Many targets for object detection are present, such as cars, ships and humans. Airplane detection is an important application performed through the use of satellites. The idea of airplane detection is to identify airplane locations in airports to reduce traffic jams. Several articles have applied various mechanisms through satellites to enable airplane detection at airports, as highlighted in this section. One of the most effective method is the application of DL techniques. This section introduces a brief background on the application of learning techniques, neural networks and deep learning methods. A literature review of related works in airplane detection using deep learning approaches is also included.

### A. AIRPLANE DETECTION

The rapid enhancements in image resolutions collected by remote sensing tools have made acquisition methods more effective. The high resolution images facilitate effective object detection. Additional data can be evaluated, allowing the detection of numerous features to increase detection efficiency. More accurate data means more detailed examples will be used for deep learning models, leading to further enhancements in the learning process. Airplanes are the most important objects in both military and civil fields; there is considerable research on aviation control and aeronautics. Most research have used satellites as the data feeder. This article addresses airport traffic jams through airplane detection using aerial images collected by drones and processed via deep learning approaches. This process is then utilised to determine the airplane type. Although several advanced technologies offer high resolution images, there are some uncertainties regarding the airplane type and position, making airplane detection an open research area. Figure 1 presents the airplane detection process. First, a remote sensing tool, such as a drone or satellite, is used to acquire the images. Next, a pre-processing action for the acquired images must be accomplished to prepare them for the detection process. Finally, the detection process is performed to identify the airplanes from the aerial images. In this method, the characteristics of the airplane would be identified automatically by using examples faded into the deep learning model. The neural network structure inside would keep updating its weights according to these examples and at the end, the model will have the ability to distinguish airplane objects inside many different images.

### B. LEARNING TECHNIQUES

Learning operations in the neural network adjust the neuron's weights of hidden layers. The neural network learns from information examples without being programmed. These examples are used as pairs, and each pair consists of an input and desired output during the learning process. The actual and desired outputs are used to calculate the precision accuracy.

Unsupervised learning techniques do not use labelled data, thus saving processing time. They apply clustering as a method for grouping samples according to how close they are to each other. The k-means clustering and Restricted Boltzmann Machines (RBMs) are well known techniques [2]. In k-means clustering, a k-number of centroids is randomly created with multiple iterations to determine the best ones. Any data close to the particular cluster centres,

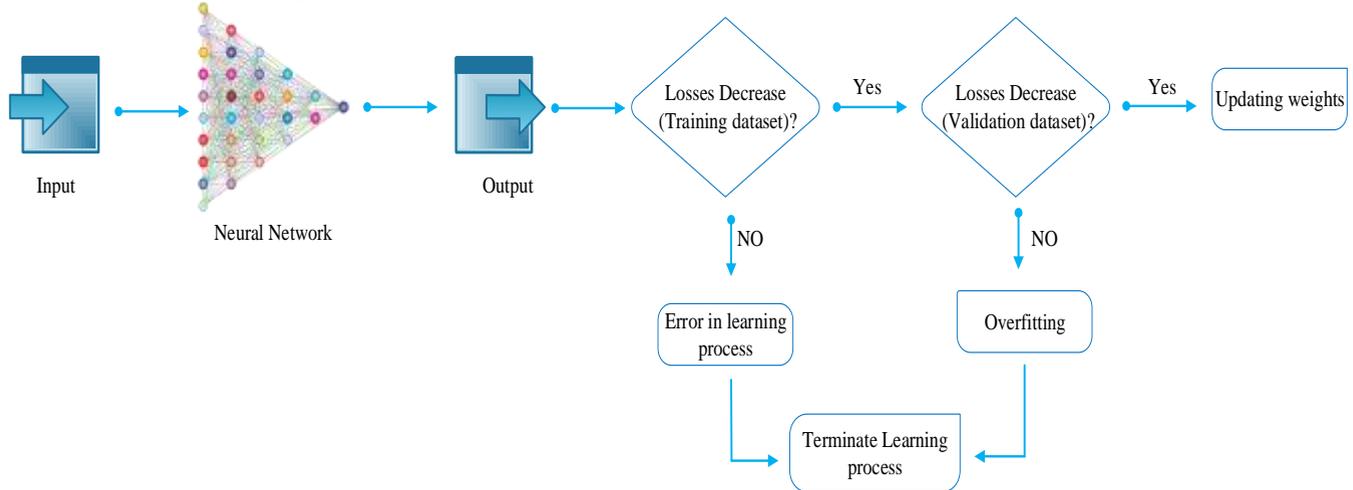

Figure 2: Concept of learning techniques

or centroids, will be assigned to that cluster.

Several applications in remote sensing (RS), such as airplane detection, employ these techniques by using the Deep Belief Network (DBN) constructed by RBMs stacking [3]. Aircraft and car detection using the k-means clustering method through the Scale-Invariant Feature Transform (SIFT) has been documented in [4]. Unsupervised convolution neural network has also been employed for extracting spectral and spatial features [5]. Ship detection using optical satellite images with a deep auto encoder structure has been proposed in [6]. The supervised learning techniques utilise training over manually labelled data, which is the main reason that enables supervised methods to obtain accurate results in the estimation of the new sample. The Support Vector Machine (SVM) is the most popular technique. It is used for classifying tasks through the development of a function that separates sample data into different categories. SVM defines a hyperplane between feature vectors extracted from samples and is considered as a margin that must be maximised.

Currently, the convolution neural network (CNN) is the appealing supervised technique. It is a neural network that works with two-dimensional image data. The convolution process that occurs inside is the main reason for its name. The convolution is a linear operation that implies the multiplication of weights with input. These weights are usually called a filter or kernel.

Filters are always smaller in size than the input. The dot product process occurs between the filter and image. After multiplication, a summation process takes place to produce a single value. The filter goes through the entire image as a sliding window; thus, it discovers the features available in an image.

Different applications are used in this field, such as object detection in RS with Deep Network (HDNN) [7]. SVM has also been applied to detect airports and airplanes [8]. In ship detection, SIFT has been employed to extract and feed SVM using panchromatic images [9]. Other researchers, such as Yang et al., used CNN to process aircraft and vehicle detection based on satellite images. The negative samples generated in RCNN affect the model detection precision, therefore, the selective research has been replaced with a saliency algorithm to avoid negative samples. The saliency algorithm uses the human visual attention mechanism to achieve bottom-up object detection [10]. Tang et al. considered two problems in Faster RCNN. The RPN layer exhibited poor performance for small-sized object localisation . Moreover the classifier is not reliable enough to distinguish vehicles and backgrounds, therefore, a hyper region proposal network (RPN) was employed to replace the classifier with a cascade of boosted classifiers [11]. Zhang et al. and Liu et al. introduced a fast region-based convolutional neural network (RCNN) method to detect ships from high-resolution remote sensing imagery [12, 13]. Alganci et al. had employed a dataset collected from public satellite sources with Single Shot Detector (SSD), faster RCNN and YOLO object detection models. The results indicate proper detection for faster RCNN and YOLO models according to COCO metrics [14]. In Traory's research, deep learning methods based on high-resolution satellite images have been used. Two types of object detection models (SSD and faster RCNN) were applied. The dataset was collected from public satellite sources such as WPU-RESISC45, WHURS19 and aerial image datasets. The results showed proper detection for SSD models according to COCO metrics[15]. Khan et al. had utilised Edge Boxes to produce proposals. These proposals are filtered by geometric checking. CNN is then used for feature extraction and classification [16].

Figure 2 presents the concept behind the learning process. The input is forwarded to the deep learning model which includes a neural network on the side. The resulting output is then compared with the desired one to observe the improvement in terms of reduction loss.

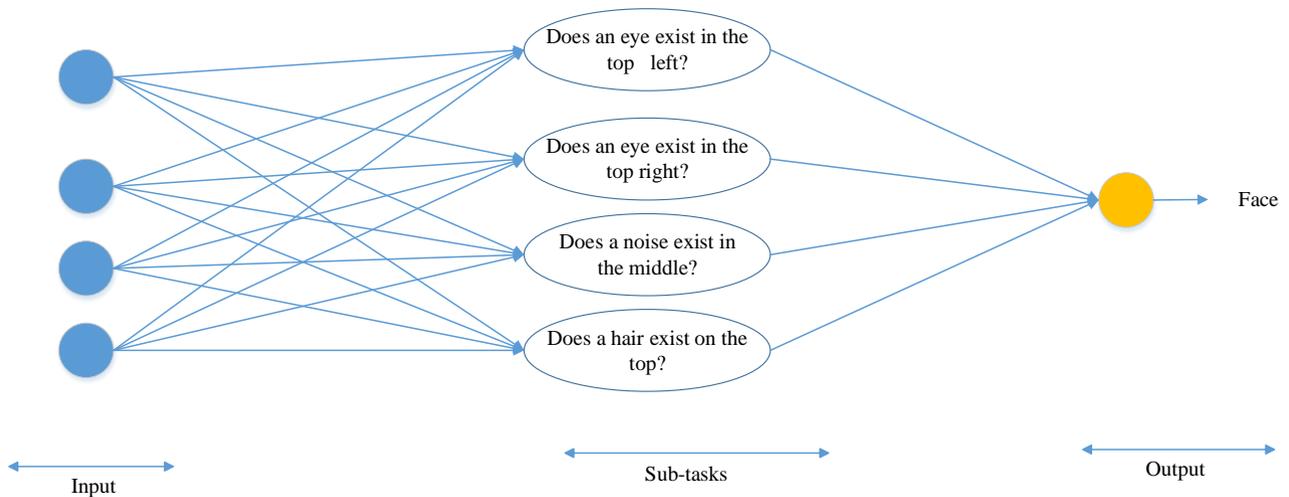

**Figure 3:** Deep learning process

If no improvement is present for multiple iterations, the learning process would then be terminated. If an improvement is detected, another comparison will be made until the last output value is performed. If there is an improvement, weights must be updated, otherwise overfitting may occur, and the learning process will be terminated.

*C. DEEP LEARNING*

The neural network performance is fantastic, but how it achieves its goals remain a mystery. It can be randomly initialised and trained by many samples, then subsequently recognise images that it has never seen before! There are no concepts of weights or biased values since these values are discovered automatically.

The main goal of the neural network is to discover the mind algorithm. This explains why inventors working on neural networks have developed artificial intelligence: to understand how the mind works. Unfortunately, that adds another challenge: how do neural networks operate?

The human face detection task can be accomplished without directly using a learning algorithm or neural network. The problem can be divided into sub-problems such as: does the image have an eye at the top left or at the top right? Is the mouth at the bottom of the image? Is the nose in the middle? If the answer to most of these sub problems is yes, then this image is probably a human face image, as shown in Figure 3. Since neural networks can address these sub problems, it is therefore better to assign a neural network to the human face detection task. The sub problems are not as simple as they seem to be. They must be decomposed into more straightforward and simpler sub problems. For example, for an eye on the top left, is there an iris, eyebrow, etc… until they reach the pixel scale. Accomplishing these tasks for numerous series of layers signify the "Deep Neural Network" [17].

*D. CONVOLUTIONAL NEURAL NETWORK (CNN)*

Convolutional Neural Network (CNN) works with two-dimensional image data. The convolution process that occurs within is the main reason for its name. Convolution is a linear operation that implies the multiplication of weights with input. These weights are usually called filters or kernels. The filters are always smaller in size than the input, and multiplication takes place in the dot product. After multiplication, a summation process is accomplished to produce a single value. This filter goes through the entire image as a sliding window. It can discover the features anywhere in an image. The workflow of the convolution process used in CNN can be simplified by illustrating edge detector using $3 \times 3$ filters with weights chosen specifically to detect vertical edges for a grayscale image from light to dark, as shown in Figure 4.

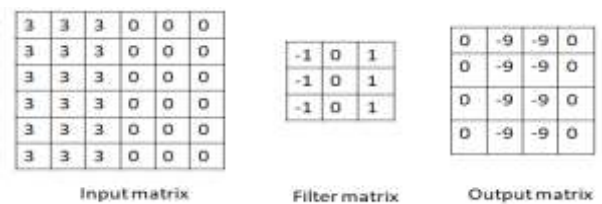

**Figure 4:** Stride size =1

The convolution process results in a smaller width and height than the original input. Feeding the convolutional layer output to another one consequently causes size shrinkage. The neural networks will then be limited in the number of layers if there is no treatment for this shrinkage. A popular treatment is to pad the entire image with enough zeroes such that the output shape will have the same width and height as the input. This is called the SAME padding. Leaving convolution without treating its shrinking effect is called valid padding, as illustrated in Figure 5.

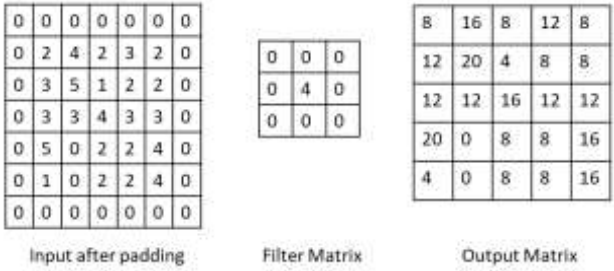

**Figure 5:** Padding effect

The stride refers to the number of pixels jumped each time by the filter in the width and height dimensions for scanning to compute the dot-product, as shown in Figure 6.

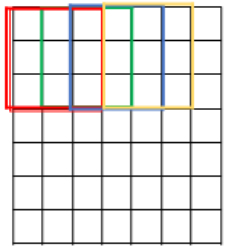

**Figure 6:** Vertical edge detector

This leads to the extension of the filter's depth to match the depth dimension of the input. For example, if the input depth is two, then the filter should match the same depth. The point is that the increased number of values should be added for each output point.

Output depth may be controlled by stacking multiple convolutional filters. While computing its results, each filter acts independently. All results are stacked together to create the output. Therefore, adding or removing convolutional filters leads to the control of the output depth.

The generalised dimensions of output layer for each input of $n \times n \times dn$, filter of $f \times f \times df$, padding p and stride $s$ are shown in Eq. 1:

$$Output\ dim = \frac{n+2p-f}{s}+1 \times \frac{n+2p-f}{s}+1 \times \#filters \quad (1)$$

### E. STATE-OF-THE-ART METHODS IN OBJECT DETECTION

*Region-based Convolutional Neural Networks (R-CNN):* This method uses an object proposal algorithm, called Selective Search, which reduces the number of bounding boxes that are fed to the classifier; approximately 2000 region proposals. The selective search employs local cues such as texture, intensity and colour to generate all possible locations of an object. These boxes can be introduced to the CNN based classifier. Since the fully connected part of CNN takes a fixed-sized input, all generated boxes should have a fixed size (224×224 for VGG) and fed to the CNN part.

*Fast RCNN*: This technique directly inputs images into ConvNet, which in turn, generates RoI. As previously mentioned, fully connected layers require fixed-size inputs. The pooling layer is applied on RoI (RoI Pooling layer) to prepare it to pass through a fully connected network (the softmax layer) which determines classes, and then the regression layer which determines boxes.

*Faster RCNN*: This method replaces selective search with a tiny convolutional network, called Region Proposal Network (RPN), to generate regions of Interest (RoI). It introduces the idea of anchor boxes with different scales (128x 128, 256×256 and 512×512) and different aspect ratios (1:1, 2:1 and 1:2) to handle the variations in both the aspect ratio and scale of objects.

*Region-based Fully Convolution Networks (RFCN):* This model has the ability to predict bounding boxes and classify them inside images. It contains a full convolution structure. It has lower computing ability per bounding box compared to faster RCNN

*Mask RCNN*: This technique extends faster R-CNN by adding a branch for predicting segmentation masks on each Region of Interest (RoI) and replaces RoIPool with RoIAlign. RoIAlign does not adjust the input proposal to correctly fit the feature map. It merely takes the object proposal and divides it into four equal bins using the bilinear interpolation: top left, top right, bottom left and bottom right. It then applies the average or maximum function to obtain the value.

*YOLO (You Only Look Once):* This approach divides the image into a grid of M x M. Each grid predicts N bounding boxes with confidence. Confidence indicates the accuracy of the bounding boxes and simultaneously determines whether or not an object is contained inside for each class in training.

*SSD (Single Shot Detector):* This detector has the ability to detect objects that possess different scales and are on different layers. It runs a convolutional network on input image only once and calculates a feature map.

Figure 7 presents high-level graphical representations and relations between these models.

### F. RELATED WORKS

This section introduces the works related to object detection used to detect airplanes, ships, vehicles..etc. It is not just for airplanes since the airplane is just one object of many other objects that could be detected using the same methods with some enhancements. Alganci et al. had used a dataset collected from public satellite sources with Single Shot Detector (SSD), faster RCNN and YOLO object detection models. The results exhibited encouraging detection capability for faster RCNN and YOLO models according to COCO metrics [14]. Zhang et al. and Liu et al. introduced a fast region-based convolutional neural network (R-CNN) method to detect ships from high-resolution remote sensing imagery [12][13].

Liu et al. worked on two problems in Faster RCNN: a region proposal network (RPN) layer which exhibited poor performance for small-sized object localisations and a classifier which is not reliable enough to distinguish vehicles and backgrounds. A hyper region proposal network (HRPN) was employed to replace the classifier by a cascade of boosted classifiers [13]. Yang et al. utilised a convolution neural network (CNN) on satellite images for aircraft and vehicle detection. The selective search was replaced with saliency due to several negative samples in the generated region proposal for RCNN, affecting the model's detection precision. Saliency uses the human visual attention mechanism to achieve the bottom-up object detection [10]. Bi et al. examined ship detection using scale-invariant feature transform (SIFT) previously extracted to feed a support vector machine (SVM) with the use of panchromatic images [9].

Li et al. proposed a robust airplane detection method in satellite images. The proposed approach is based on saliency computation and symmetry detection. This method stably performs in acquiring locations and orientations [18]. Sun et al. offered a detection framework based on spatial sparse coding bag-of-words to solve target detection for complex shapes in high–resolution remote sensing images. The suggested framework was combined with a linear support vector machine for target detection [19]. Cai et al. focused on aircraft automatic segmentation from high-resolution satellite images based on the idea of co-segmentation. They first selectively segmented out the regions of interest, then applied a region-based shadow detection to remove shadows. Finally, anisotropic heat diffusion was employed to fulfil co-segmentation [20]. Liu et al. proposed a coarse-to-fine process. In the first stage, the position of the airplane was roughly estimated using single template matching. In the second process, a parametric shape model using principal component analysis was accomplished. In the final step, parameters of the segmentation results were directly applied to verify the type with two k-nearest neighbour steps [21]. Hsieh et al. proposed a hierarchical classification algorithm for aircraft recognition in satellite images. The aircraft object appears on a small scale in satellite images, in addition to noise and dazzle pints which represent challenges. The paper proposed a novel booting algorithm to learn a set of proper weights from training samples for feature integration [22]. Dai, He and Sun presented a multi-task network cascade for instance-aware semantic segmentation. The model consists of three networks that differentiate instances, estimate masks and categorise objects. They developed an algorithm for nontrivial end-to-end training of the casual cascaded structure [23]. Cao et al. introduced a novel content-based remote sensing image retrieval method based on the triplet deep metric learning convolutional neural network. The extracted representative features are in semantic space so that the images from the same class are close to each other. Those that are from different classes are far apart [24]. Filippidis, Jain & Martin used fuzzy reasoning to enhance the accuracy of the automatic detection of aircrafts in synthetic aperture radar images. Prior knowledge obtained from colour aerial photographs were used to achieve a noticeable enhancement in false alarm [25]. Marmanis et al. introduced a novel method consisting of a two-stage framework. The first stage is a pre-trained CNN designed for tracking an entirely different classification problem, then exploiting it to extract an initial set of representations. In the second stage, the driven representations are transferred into a supervised CNN classifier [26]. Castelluccio et al. prepared a Land Use Classification in Remote Sensing Images by Convolutional Neural Networks [27]. Ruiz, Fdez-Sarría & Recio analysed texture methods applied to the classification of remote sensing. They created a database with high resolution in both satellite and area images. The texture classes were defined in urban applications. In the first application, it is crucial to know the type of vegetations. A comparative study was also conducted on texture feature extraction for the classification of remote sensing data using wavelet decomposition [28]. In [29], the classification of satellite images with regularised AdaBoosting of RBF neural networks was accomplished. In [30], the authors introduced a sparse coding method for satellite scene classification. They further presented local ternary pattern histogram Fourier features and combined a set of divers with complementary features to enhance performance. A high-resolution satellite scene classification using a sparse coding-based multiple feature combination was also created. Table 1 presents a summary of several relevant research articles in this field.

### G. THE PROBLEM

The massive number of flights places tremendous pressure on airports. Some airports may have the capacity to accommodate numerous daily flights. However, in future, they will most likely experience traffic jams. Based on the statistics provided in this paper, the frequency of flights is expected to dramatically increase. Although the covid-19 pandemic has critically affected last year's number of flights, they are beginning to steadily increase. It has been previously determined that the best methods for addressing airport traffic jams are deep learning approaches using satellite images. Unfortunately, satellite technology is very costly and requires advanced technology. In addition, weather conditions may alter the entire process. It cannot be a universal solution since some countries cannot employ an entire satellite for just one purpose, which is to monitor an airport. This paper addresses the issue by proposing the use of aerial images acquired from drones. The aerial images are standard images with higher resolution which do not require pre-processing in contrast to satellite images. In addition to the detection function, this paper prepares future research for identifying the type of each detected airplane based on information collected by the estimated masks.

**TABLE 1** SUMMARY OF PREVIOUS CONTRIBUTIONS

| No | Author | Contribution |
|---|---|---|
| 1 | Alganci, Soydas & Sertel | SSD, faster RCNN and YOLO to detect airplanes using satellite images [14] |
| 2 | Zhang et al. | RCNN for ships detection using high-resolution images [12] |
| 3 | Lei et al. | Employed an HRPN and replaced the classifier by a cascade of boosted classifiers in faster R-CNN [13] |
| 4 | Yang et al. | CNN on satellite images to conduct aircraft and vehicle detection [10] |
| 5 | Bi et al. | Ship detection using SIFT extracted previously to feed an SVM using panchromatic images [9] |
| 6 | Cheng & Han | Rcnet, SPMK, SSC, SSAE for object detection [31] |
| 7 | Xiang et al. | HDNN for vehicle detection [7] |
| 8 | Xiang et al. | DBN for aircraft detection [3] |
| 9 | Chen, Zhan & Zhang | SSD for object detection [32] |
| 10 | Yang et al. | R-DFPN- Fast-RCNN-Resnet-101 for ships detection [33] |
| 11 | Mundhenk et al. | YOLO for car detection [34] |
| 12 | Zhou et al. | A framework designed for weakly supervised target detection in R based on transferred deep features and negative bootstrapping [35] |
| 13 | Gardner & Dorling | A review of applications in the atmospheric sciences [36] |
| 14 | Liu et al. | Method for object detection in images named single shot multibox detector (SSD) [37] |
| 15 | Castelluccio et al. | Semantic segmentation for remote sensing scenes using two datasets: CaffeNet and GoogleLeNet [27] |
| 16 | Li et al. | A pixel-pair model proposed to achieve similarities between pixels [38] |
| 17 | Mou, Ghamisi & Zhu | A fully conv-deconv network with residual learning for unsupervised spectral-spatial feature extraction [39, 40] |
| 18 | Sheng et al. | A classification method for satellite images using sparse coding [30] |
| 19 | Fan, Chen & Lu | Multiple spare coding architecture to capture multiple aspects of discriminative structure [41] |
| 20 | Dai & Yang | Image classification via two-layer sparse coding with promising results without using the learning phase [42] |
| 21 | Camps-Valls & Rodrigo-González | Classification of satellite images with regularised AdaBoosting of RBF neural networks [29] |
| 22 | Ruiz, Fdez-Sarría & Recio | Analyses several texture methods applied to the image classification [28, 43] |
| 23 | Li & Castelli | Derives texture features set for content-based retrieval of satellite image database [44] |
| 24 | Bruzzone & Carlin | Pixel-based system for geometrical spatial images [45] |
| 25 | Zhao & Principe | Real application of SVM for SAR [46, 47] |
| 26 | Deng et al. | New dataset called large-scale hierarchical image database (ImageNet) [48] |
| 27 | Shin et al. | Explores and evaluates distinct CNN architectures (5 thousands-160 million) parameters. The evaluation process is accomplished to examine when and why transfer learning can be useful [49] |
| 28 | Yang et al. | A multiscale rotational region detection approach was introduced. The approach has the ability to deal with various complex senses with many different objects. It also has the ability to reduce redundant detection regions [33]. |
| 29 | Liu et al. | A coarse-to-fine aircraft recognition approach where similarities and differences in shapes are explored. Single templates with a defined score were adopted for the estimation [21]. |
| 30 | Hsieh et al. | A classification approach for aircraft recognition in satellite images with high accuracy [22] |
| 31 | Moldovan & Wu | A robust arrangement for the knowledge of object-level and meta-level in conjunction with the knowledge of the domain. The purpose is to facilitate fast interference in real-time by using one or multi processors [50]. |
| 32 | Pesaresi & Benediktsson | A segmentation approach of connected components in images based on morphological characteristics [51] |
| 33 | Greenberg & Guterman | Description of multilayer perceptron network and adaptive resonance theory to UAV recognition problem [52] |
| 34 | Abadi et al. | TensorFlow is a machine learning framework with a large scale and heterogeneous environment. Its architecture provides flexibility to the developers to experiment with their optimisations and algorithms [53]. |
| 35 | Redmon & Farhadi | Some updates and enhancements have been added to YOLO for improvement and accuracy [54] |
| 36 | Wojek et al. | Presented a fast object class localisation approach implemented on a data-parallel architecture. The implementation of HOG showed that the speed of the CPU increased dramatically [55]. |
| 37 | Nie et al. | The first study in ship detection in satellite images using SSD [56] |
| 38 | Radovic, Adarkwa & Wang | Provided the parameter selection procedure to train a CNN. Selecting the suitable set of training parameters would enhance the results in terms of classification and detection. Objects may be tracked and classified in a video captured by a drone using the YOLO model [57]. |
| 39 | Ammour et al. | Presented an automatic approach to address detecting and counting cars using UAV. The images must be high resolution so the segmentation process can be applicable in high efficiency. The detection locations for each car can be determined by segmenting each input image into small homogeneous areas [58]. |
| 40 | Li et al. | Faster RCNN, YOLO and SSD were used for detecting agricultural greenhouses (AGs). Evaluation results indicate that YOLO has the best performance based on average precision mAP [59]. |
| 41 | Reda & Kedzierski | A faster edge region CNN algorithm was proposed to enhance the accuracy of building detection and classification [60]. |
| 42 | Hong et al. | Four models, faster RCNN, region-based fully CNN, YOLO and SDD, were used to conduct bird detection tasks. Results showed that Faster RCNN is the most accurate and YOLO is the fastest [61]. |
| 43 | Nguyen et al. | Evaluation for several deep learning models, such as faster RCNN, region-based fully CNN, SSD and YOLO, based on object detection for videos captured by a drone [62] |
| 44 | Zhang et al. | A depth-wise separable operation convolution method was performed to optimise the convolutional layer of tiny YOLO. The aim is to reduce the calculation burden and improve speed [63]. |

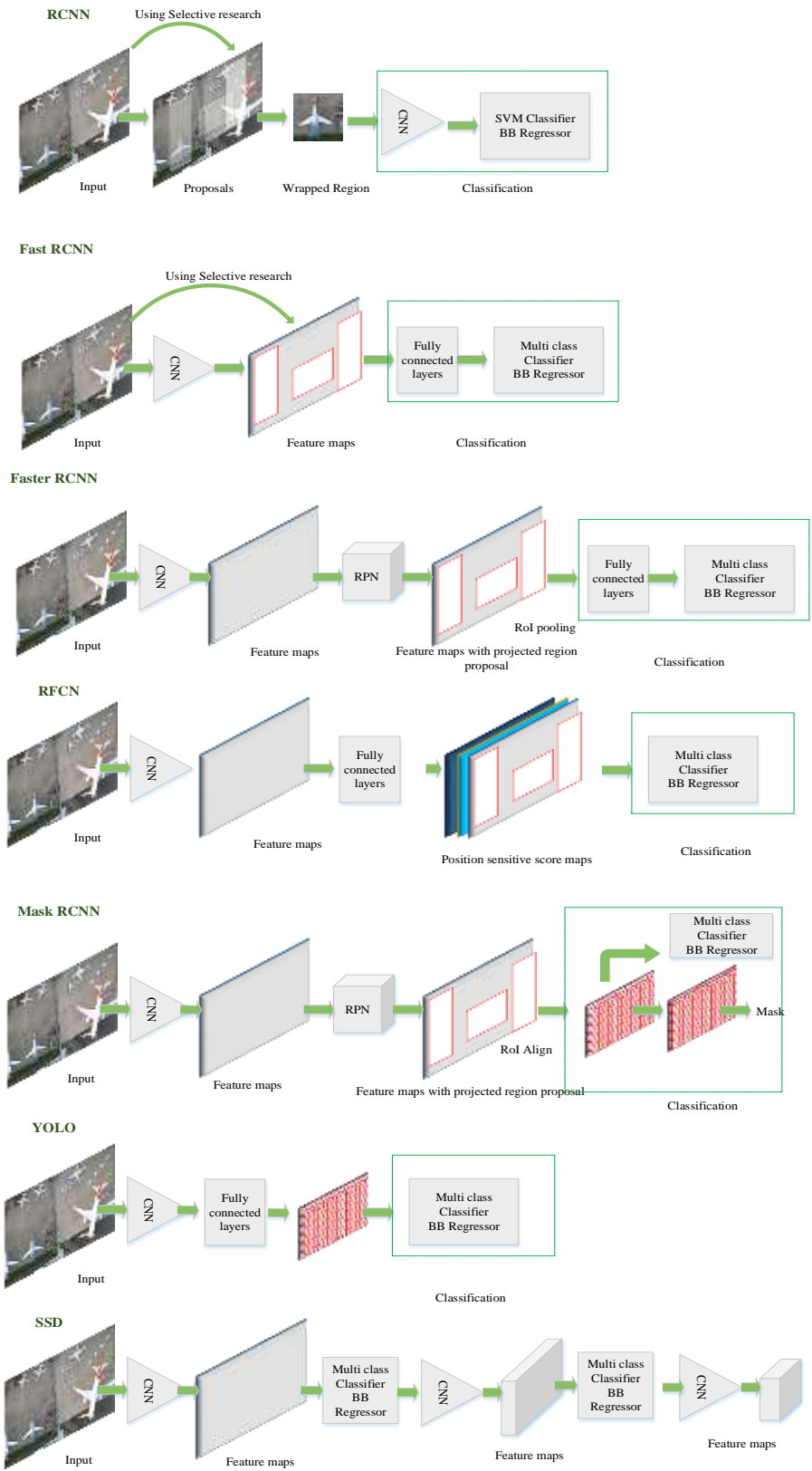

**Figure 7**: High-level graphical representations and relations between various state-of-the-art methods in object detection

## III. SYSTEM AND SIMULATION MODEL

This section begins with drone image acquisition. The images were collected by two eBee Classic drones flying simultaneously for Le Bourget airport in Paris[64]. These images have a ground resolution of 3.14 cm/px with 4608 x 3456 px dimensions. This resolution is difficult to process with the computational environment used in the study, as presented in Table 2.

**TABLE 2** COMPUTATIONAL ENVIRONMENT

| Component name | Specifications |
|---|---|
| Processor | Intel(R) Core(TM) i7-7500 U CPU |
| Speed | 2.70 GHz – 2.9 GHz |
| RAM | 8.00 GB |
| Graphics Components and Specifications | |
| Chip Type | GeForce 940MX |
| Memory | 6030 MB |

### A. DATA COLLECTION

Two categorisations, in terms of lowered dimensional size, have been accomplished to simplify the image processing. The first category is the size of 369 x 259 px (considered as low resolution) and has the smoothest processing mode. The second category is the size of 922 x 864 px (considered as high resolution) and has a slower processing mode. Some images have no airplane object inside and some have been repeated more than once. Thus, a manual filtration has been accomplished, adding another classification to the low resolution categorisation. The categorisations of the dataset are presented in Table 3.

**TABLE 3** DATASET CLASSIFICATION

| Dataset Categorisations | | # |
|---|---|---|
| High-resolution General | | 557 |
| Low resolution | General | 557 |
| | Airplane Objects | 94 |

Data augmentation processing has been used to enhance the learning proficiency later on. The augmentation varies between noise addition, rotation and cropping for each classification in the dataset to increase the number of images, as illustrated in Table 4.

This paper has applied a supervised technique which uses manually labelled examples for training. "LabelImg"[65] and the "PixelAnnotationTool" [66] have been utilised to label data within bounding boxes and semantic segmentation, as shown in Figures 8 and 9, respectively.

**TABLE 4** DATASET AFTER AUGMENTATION

| Dataset Categorisations | | # |
|---|---|---|
| High-resolution General | | 1114 |
| Low resolution | General | 1114 |
| | Airplane Objects | 689 |

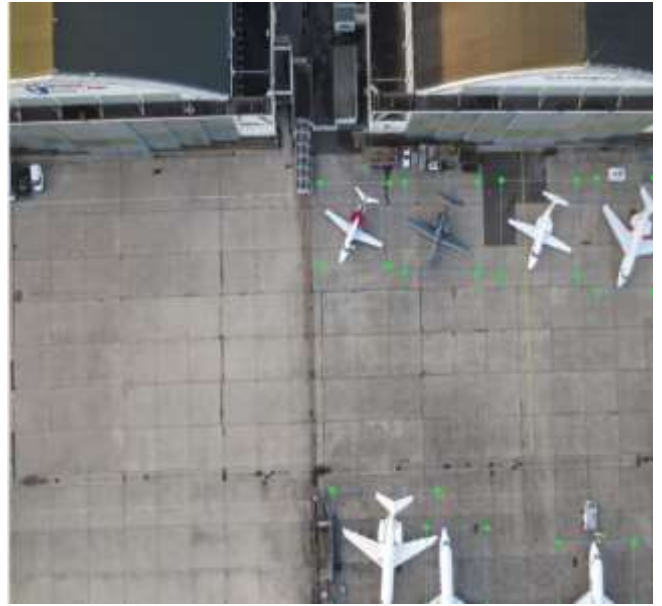

**Figure 8**: Bounded boxes labelling

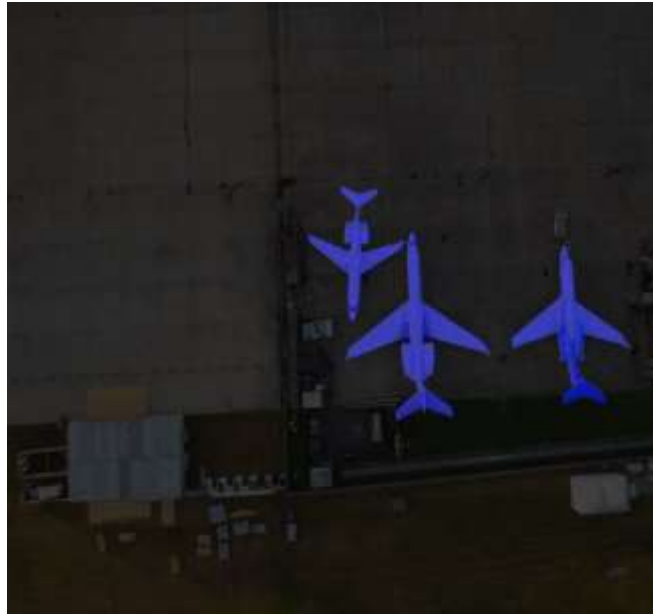

**Figure 9**: Semantic segmentation

**TABLE 5** MODELS

| | Model 2 | Model 3 | | Model 4 | Model 5 | Model 6 | Model 7 | Model 8 |
|---|---|---|---|---|---|---|---|---|
| Dataset | Low Resolution with Airplane | Low Resolution with Airplane Object | | Low Resolution with Airplane Object | Low Resolution with Airplane Object | Low Resolution with Airplane Object | High Resolution | Low Resolution |
| Additional changes configurations on | Image resize (230 x 350) | first_stage_features_stride | 8 | Learning rate Manual_Step_Learning_Rate = 0.002 Schedule= 0.0002, 0.00002 | first_stage_max_proposals 100 | Learning rate Manual_Step_Learning_Rate = 0.00002 Schedule= 0.000002, 0.0000002 | - | - |
| | | height_stride | 8 | | | | | |
| | | width_stride | 8 | | | | | |

## B. TRAINING

The Mask RCNN model has been selected since the goal of the training model in this research is airplane detection. This will facilitate the identification of airplane type in future works related to instance segmentation [67]. TensorFlow is a useful platform for deep learning as it provides pre-trained models that help initialise neural networks [53]. The mask_rcnn_inception_v2_coco configuration has been applied from the TensorFlow models. It is the base configuration for the training process utilised in this research. The configurations of hyperparameters have no optimal values and their values change as datasets change [17]. Thus, different changes have been made on the base configuration. The changes are categorised into 7 additional models, as shown in Table 5. Figure 10 presents the models' training performance by TensorBoard [68]. The results indicate that the loss decreases for all models as several steps increase.

A continuous assessment of the validation set has been conducted for each model. Approximately all models exhibited a steady-state as more steps increased, except for Model 3 which continued to improve.

## C. SOFTWARE

Several software programs have been applied, together with machine specifications, to ensure the enhanced performance of the machine during training processes. These software include the following: CUDA is a platform created by Nvidia to allow software developers and engineers to employ the CUDA-enabled graphics processing unit for general-purpose processing. cdDNN is a GPU-accelerated library for deep neural network training purposes. Jupyter is an open-source software and service for interactive computing across multi-programming languages. Google Colab is a free online cloud service that offers the operation of deep learning or machine learning models. Anaconda is an open-source distribution for Python and R programming languages. Its goal is to simplify package management and deployment. Python is a high-level general-purpose programming language. Atom is an open-source code editor for macOS, Linux and Microsoft Windows operating systems. TensorFlow is a well-known research platform for deep learning that can modify and implement different types of deep learning models.

## IV. RESULTS AND DISCUSSIONS

This paper has implemented an evaluation process to measure the performance of plane detection models. Several essential points must first be highlighted. Multiple metrics have been defined for object detection performances adopted by popular competitions such as PASCAL VOC, COCO and Open Images Challenges. All of them use the mean average precision as a fundamental metric; however, there are some variations in definitions. COCO employs the average recall as a new metric. The confidence score and Intersection over Union (IoU) are fundamental concepts in the evaluation process. Table 6 presents the full description of COCO metrics. The confidence score is the likelihood that the classifier's probability will find an object inside an anchor box. These two terms are used to determine whether detection is a true positive (TP), a false positive (FP), a true negative (TN) or a false negative (FN). TP occurs if three conditions have been satisfied: the confidence score is higher than the threshold, the predicted class matches a class of ground truth and IoU is greater than the threshold. However, if one of the last two conditions has not been met, FP will then occur. If a confidence score that has to detect a ground-truth is lower than the threshold, it will cause a false negative detection (FN). If a confidence score that has not detected anything is lower than the threshold, it will cause a true negative detection (TN). For analytical purposes, the terms 'precision' and 'recall' are introduced. Precision is the number of true positives divided by the sum of true positives and false positives, as shown in Eq. 2. Recall is the number of true positives divided by the sum of true positives and false negatives (also known as the number of ground-truths), as displayed in Eq. 3.

$$precision = \frac{TP}{TP + FP} \qquad (2)$$

$$recall = \frac{TP}{TP + FN} = \frac{TP}{\# \, ground \, truths} \qquad (3)$$

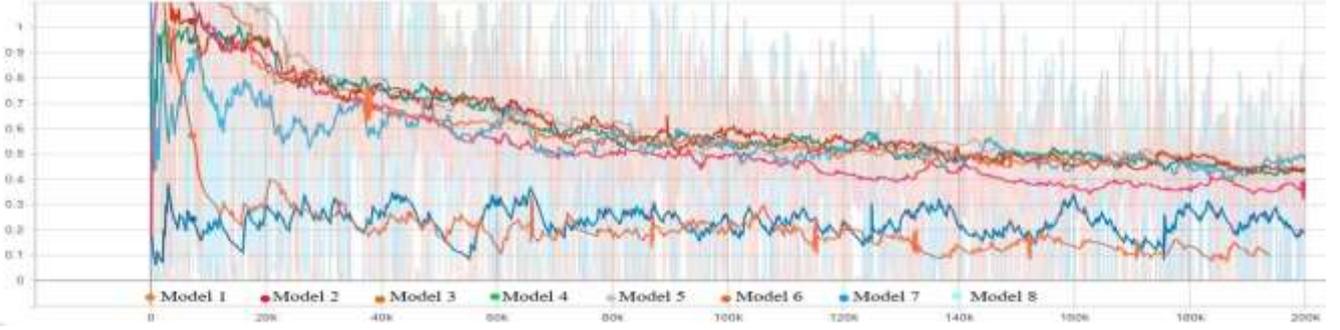
**Figure 10**: Training progress

Creating multi-variations in the threshold value of confidence score produces multiple values for both recall and precision. Mapping these values with the x-y axis forms a precision-recall curve that indicates the association between the two metrics. Another curve with a special advantage of evaluating the effectiveness of detection proposals is also present. This curve is formed by producing variations in the IoU threshold, leading to different related values of recall. Since various detectors are used in object detection, introducing other numerical metrics is necessary, such as the average precision (AP), mean average precision (mAP), average recall (AR) and mean average recall (mAR), as provided in Eqs. 4, 5, 6 and 7. AR is the average precision over all recall levels. It is preferable to reduce wiggles in the curve by using interpolated precision $p_{interp}$ at a certain recall level r, then calculating the area under the resultant curve. The mAP is the average of AP over all classes C. The AR is the average recall over all IoUs in the interval [0.5, 1.0].

$$p_{interp} = \max_{r' \geq r} p(r') \quad (4)$$

$$AP = \sum_{i=1}^{n-1} (r_{i+1} - r_i)\ p_{interp}(r_{i+1}) \quad (5)$$

$$mAP = \frac{\sum_{i=1}^{C} AP_i}{C} \quad (6)$$

$$AR = 2 \int_{0.5}^{1} recall(IoU)\ dIoU \quad (7)$$

In Eq. 4, the $p(r')$ is the measured precision at recall $r'$. The precision value of recall $r'$ is replaced with the maximum precision for any recall $\geq r'$. In Eq. 5, instead of sampling a fixed number of points, $p_{interp}(r_i)$ is sampled so that it drops to compute AP as a sum of rectangles.

In this article, COCO metrics have been applied for evaluation purposes. COCO has multiple mAP metrics defined by multiple IoU thresholds. No improvement was present in inaccuracy values associated with an increase in the number of examples used in the training dataset for Model 1 versus Model 8. Normally, additional examples enhance the learning efficiency, therefore increasing accuracy. Numerous outcomes of deep learning cannot be explained due to the mystery of neural network behaviour. The clear example is that their weights randomly initialise, but exactly how they adjust to identify the desired outputs remains unknown. The neural network should be considered as a black box that has inputs and outputs with hidden layers inside. Various techniques have been utilised to optimise and handle several other hyperparameters. However, we can assign the non-enhancement to rezone the examples that do not contain airplane objects inside. There is an improvement in inaccuracy values associated with the increase of resolution of examples used in the training dataset for Models 8 versus Model 7. The metrics of small scale areas have been excluded due to the definition of small objects; small objects are those which contain areas of less than 32² pixels. The increase in resolution have increased the number of pixels.

The first challenging metric measured the AP of IOU = (0.5:0.05:0.95). Model 6 achieved the highest values of 0.921 for the training dataset and 0.573 for the test dataset.

The second metric is the PASCAL metric which measured the AP of IOU = (0.5). All models have satisfied a high value of 0.99 for the training dataset, except Models 3 and 8. Only Model 7 achieved a high value of 0.955 for the test dataset.

The third metric is the strict metric which measured the AP of IOU = (0.75). Models 4, 5 and 7 achieved a value of 0.99 for the training dataset, however, only Model 5 attained the highest value of 0.652 for the test dataset.

The fourth metric measured the AP for small objects that have been defined as areas smaller than 32² pixels. Models 4 and 6 achieved the highest values of 0.875 for the training dataset, and only Model 6 attained the highest value of 0.426 for the test dataset.

The fifth metric measured the AP for medium objects that have been defined as objects which occupy areas in the range of 32² and 96² pixels. Only Model 5 obtained the highest values of 0.943 for the training dataset and 0.628 for the test dataset.

The sixth metric measured the AP for large objects that have been defined as objects which occupy areas larger than 96². Only Model 2 achieved the highest values of 0.978 for the training dataset and 0.808 for the test dataset.

The seventh metric measured the AR of one detection per image. Model 7 acquired the highest value of 0.434 for the training dataset, however, Model 6 achieved the highest value of 0.289 for the test dataset.

The eighth metric measured the AR of ten detections per image. Model 6 attained the highest values of 0.942 for the training dataset and 0.617 for the test dataset.

TABLE 6 DESCRIPTION OF COCO METRICS

| Metrics | IoU | Description | Number of metrics |
|---|---|---|---|
| mAP | 0.50:0.5:0.95 | 10 IoU thresholds (i.e., 0.50, 0.55, 0.60, …, 0.95) | 10 |
| mAP | 0.50 | Pascal VOC metric | 1 |
| mAP | 0.75 | strict metric | 1 |
| | | Total # metrics | **12** |
| **Metrics** | **IoU** | **Description** | **Number of metrics** |
| mAP Small | 0.50:0.5:0.95 | objects with areas $< 32^2$ | 10 |
| mAP Medium | 0.50:0.5:0.95 | $32^2 <$ objects with areas $< 96^2$ | 10 |
| mAP Large | 0.50:0.5:0.95 | objects with areas $> 96^2$ | 10 |
| | | Total # metrics | **30** |
| **Metrics** | **max** | **Description** | **Number of metrics** |
| mAR | 1 | Max 1 object / image | 1 |
| mAR | 10 | Max 10 objects / image | 1 |
| mAR | 100 | Max 100 objects / image | 1 |
| | | Total # metrics | **3** |
| **Metrics** | **area** | **Description** | **Number of metrics** |
| mAR | Small | objects with areas $< 32^2$ | 1 |
| mAR | Medium | $32^2 <$ objects with areas $< 96^2$ | 1 |
| mAR | Large | objects with areas $> 96^2$ | 1 |
| | | Total # metrics | **3** |

The ninth metric measured the AR of a hundred detections per image. Model 6 achieved the highest values of 0.943 for the training dataset and 0.625 for the test dataset.

The tenth metric measured the AR for small objects that have been defined according to areas smaller than 32² pixels. Model 4 obtained the highest value of 0.908 for the training dataset, however, Model 6 reached the highest value of 0.504 for the test dataset.

The eleventh metric measured the AR for medium objects which have been defined as objects that occupy areas in the range of 32² and 96² pixels. Only Model 5 acquired the highest values of 0.963 for the training dataset and 0.677 for the test dataset.

The twelfth metric measured the AP for large objects which have been defined as objects that occupy areas larger than 96² pixels. Model 2 achieved the highest value of 0.985 for the training dataset, however, Models 5 and 6 attained the highest value of 0.829 in the test dataset.

In conclusion, Model 6 exhibited the best performance among all other models. This was expected owing to its lower learning rate. Model 5 displayed perfect results due to a decrease in the max proposal regions of 100. In general, Tables 6 and 7 offer promising values in terms of COCO metrics in comparison to other research results that had implemented satellite images. Table 8 is the benchmark which presents the detection accuracy for all eight models on satellite images. 200 images have been randomly taken from airplane categorisations of the NWPU-RESISC45 dataset. Although satellite images are different from drone images, the accuracy values seem to be similar to other previous results, indicating the reliability of using drone images with deep learning algorithms for airplane detection.

The values shown in the tables further signify the reliability of the approach employed in this paper. COCO metrics are one of the best metrics used to determine the accuracy of applied methods in object detection. It can identify whether the utilised method is better than others according to metric values. Although this approach employed drone images instead of satellite images, the resulting values are more promising since they had mostly achieved the same accuracies or higher than what was previously stated in the literature review.

As mentioned in the introduction, the goal of this research is to provide a universal solution for managing traffic jams at airports. The proposed approach can be applied in any country around the world since it does not require significant cost or complicated technology as compared to other solutions. Another advantage is the fast pre-processing of images which may not need further pre-processing since drone images are standard images. Future endeavours should focus on identifying the airplane type according to its length. The proposed idea considers the advantages of instance segmentation of each detected plane, together with the drone's high-resolution images, to calculate the surface area and then determine the length. Length can be considered as a key factor to establish the detected airplane type.

TABLE 7 TRAINING DATASET EVALUATION

| Model | Metrics | | | | | | | | | | | |
|---|---|---|---|---|---|---|---|---|---|---|---|---|
| | 1 | 2 | 3 | 4 | 5 | 6 | 7 | 8 | 9 | 10 | 11 | 12 |
| 1 | 0.899 | **0.99** | 0.98 | 0.848 | 0.917 | 0.931 | 0.369 | 0.927 | 0.927 | 0.882 | 0.947 | 0.952 |
| 2 | 0.898 | **0.99** | 0.98 | 0.819 | 0.918 | **0.978** | 0.399 | 0.918 | 0.918 | 0.849 | 0.943 | **0.985** |
| 3 | 0.269 | 0.577 | 0.205 | 0.053 | 0.311 | 0.688 | 0.197 | 0.348 | 0.389 | 0.157 | 0.448 | 0.755 |
| 4 | 0.892 | **0.99** | **0.99** | **0.875** | 0.903 | 0.903 | 0.389 | 0.924 | 0.924 | **0.908** | 0.931 | 0.925 |
| 5 | 0.92 | **0.99** | **0.99** | 0.863 | **0.943** | 0.961 | 0.403 | 0.940 | 0.941 | 0.888 | **0.963** | 0.977 |
| 6 | **0.921** | **0.99** | 0.983 | **0.875** | 0.937 | 0.969 | 0.402 | **0.942** | **0.943** | 0.901 | 0.958 | 0.982 |
| 7 | 0.902 | **0.99** | **0.99** | 0.582 | 0.897 | 0.914 | **0.434** | 0.926 | 0.928 | 0.580 | 0.924 | 0.939 |
| 8 | 0.816 | 0.965 | 0.928 | 0.737 | 0.896 | 0.935 | 0.412 | 0.840 | 0.842 | 0.772 | 0.922 | 0.957 |

TABLE 8 VALIDATION DATASET EVALUATION

| Model | Metrics | | | | | | | | | | | |
|---|---|---|---|---|---|---|---|---|---|---|---|---|
| | 1 | 2 | 3 | 4 | 5 | 6 | 7 | 8 | 9 | 10 | 11 | 12 |
| 1 | 0.568 | 0.945 | 0.614 | 0.416 | 0.617 | 0.791 | 0.284 | 0.614 | 0.619 | 0.502 | 0.664 | 0.821 |
| 2 | 0.542 | 0.928 | 0.570 | 0.367 | 0.597 | **0.808** | 0.278 | 0.588 | 0.595 | 0.459 | 0.647 | 0.829 |
| 3 | 0.240 | 0.533 | 0.174 | 0.054 | 0.287 | 0.710 | 0.187 | 0.304 | 0.349 | 0.147 | 0.418 | 0.742 |
| 4 | 0.554 | 0.936 | 0.605 | 0.389 | 0.611 | 0.766 | 0.284 | 0.598 | 0.603 | 0.467 | 0.664 | 0.796 |
| 5 | 0.570 | 0.927 | **0.652** | 0.413 | **0.628** | 0.801 | 0.287 | 0.615 | 0.62 | 0.484 | **0.677** | 0.829 |
| 6 | **0.573** | 0.938 | 0.645 | **0.426** | 0.627 | 0.781 | **0.289** | **0.617** | 0.625 | **0.504** | 0.672 | 0.829 |
| 7 | 0.525 | **0.955** | 0.556 | 0.00 | 0.440 | 0.610 | 0.216 | 0.586 | 0.593 | 0.00 | 0.51 | 0.672 |
| 8 | 0.364 | 0.767 | 0.335 | 0.175 | 0.51 | 0.756 | 0.174 | 0.419 | 0.435 | 0.274 | 0.573 | 0.791 |

TABLE 9 SATELLITE EVALUATION AS A BENCHMARK

| Model | Metrics | | | | | | | | | | | |
|---|---|---|---|---|---|---|---|---|---|---|---|---|
| | 1 | 2 | 3 | 4 | 5 | 6 | 7 | 8 | 9 | 10 | 11 | 12 |
| 1 | 0.448 | 0.911 | 0.384 | 0.226 | 0.477 | 0.637 | 0.230 | 0.505 | 0.520 | 0.386 | 0.540 | 0.697 |
| 2 | 0.405 | 0.845 | 0.317 | 0.184 | 0.434 | 0.611 | 0.221 | 0.466 | 0.484 | 0.329 | 0.508 | 0.659 |
| 3 | 0.244 | 0.531 | 0.174 | 0.061 | 0.29 | 0.465 | 0.157 | 0.334 | 0.375 | 0.234 | 0.394 | 0.579 |
| 4 | 0.397 | 0.813 | 0.300 | 0.188 | 0.427 | 0.575 | 0.209 | 0.455 | 0.482 | 0.335 | 0.506 | 0.656 |
| 5 | 0.456 | 0.902 | 0.398 | 0.209 | 0.495 | 0.637 | **0.241** | 0.520 | 0.524 | 0.343 | 0.555 | 0.709 |
| 6 | **0.472** | **0.932** | **0.430** | **0.260** | **0.499** | **0.687** | 0.240 | **0.530** | **0.538** | **0.387** | **0.560** | **0.738** |
| 7 | 0.393 | 0.852 | 0.250 | 0.245 | 0.436 | 0.456 | 0.211 | 0.480 | 0.506 | 0.374 | 0.540 | 0.532 |
| 8 | 0.314 | 0.693 | 0.244 | 0.106 | 0.360 | 0.491 | 0.190 | 0.380 | 0.409 | 0.224 | 0.441 | 0.612 |

Figure 11 presents some of the visualisation results of the test dataset. Images A, B, C, E and G contain TP detections. Images D and F include both TP and FN detections. Image H contains TN detection. This figure reflects the values Stated in the tables. The accuracy is pretty good but not 100 percent; that's why rarely few objects would not be detected such as in figure 11 d.

## V. CONCLUSIONS

In this paper, an easy and low-cost approach has been provided to support airplane traffic control at airports. The first step of the suggested approach applies a deep learning model with aerial images collected by drones to detect airplanes. The next step for future efforts is to apply the provided data to first identify the airplane type according to its surface area and then its length. This approach can be a universal contribution since any country in the world can benefit from this technique. Drones can feed the system with aerial images rather than the satellite approach which requires advanced and costly technology. This article provides a clear explanation of the new approach and reviews previous works in the related research field. It can be used as a reference for future research and publications. The evaluation yielded promising outcomes in terms of COCO metrics. Although Satellite images are different from drone images, the accuracy values are encouragingly similar, further indicating the reliability of this approach.

**APPENDIX:**

For convenience, Table 10 lists a summary of the abbreviations introduced in the text.

TABLE 10 SUMMARY OF ABBREVIATIONS

| | |
|---|---|
| AP | Average Precision |
| AR | Average Recall |
| CNN | Convolutional Neural Network |
| COCO | Common Objects In Context |
| CPU | Central Processing Unit |
| CUDA | Computer Unified Device Architecture |
| CUDNN | The Nvidia Cuda Deep Neural Network Library |

| | | | | |
|---|---|---|---|---|
| DBN | Deep Belief Network | | RCNN | Region Convolution Neural Network |
| DL | Deep Learning | | ReLU | Rectified Linear Unit |
| Faster RCNN | Faster Region-Based Convolutional Network | | RNN | Recurrent Neural Network |
| | | | RPN | Region Proposal Network |
| FN | False Negative | | RS | Remote Sensing |
| FP | False Positive | | RS | Remote Sensing |
| GPU | Graphics Processing Unit | | SGD | Stochastic Gradient Descent |
| HDNN | Deep Network | | SIFT | Scale-Invariant Feature Transform |
| HRPN | Hyper Region Proposal Network | | SIFT | Scale-Invariant Feature Transform |
| IoU | Intersection over Union | | SVM | Support Vector Machine |
| mAP | Mean Average Precision | | TN | True Negative |
| mAR | Mean Average Recall | | TP | True Positive |
| RAM | Random Access Memory | | YOLO | You Only Look Once |
| RBM | Restricted Boltzmann Machines | | | |

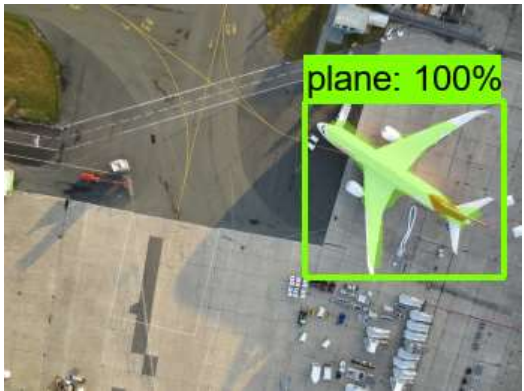

A

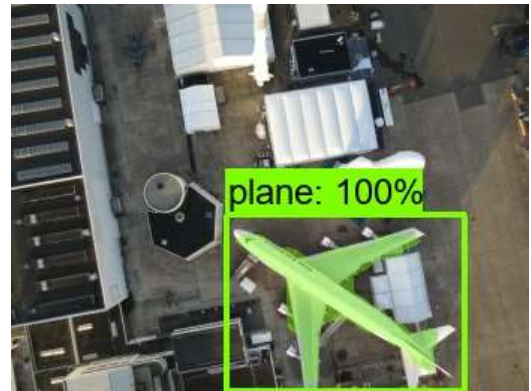

B

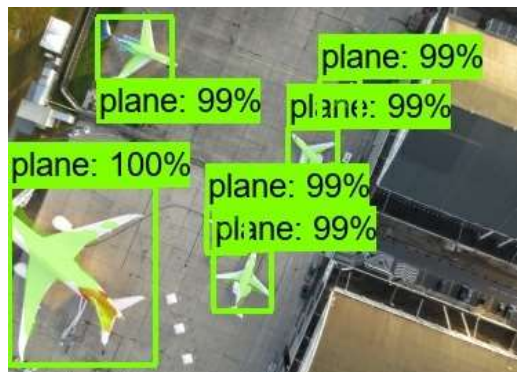

C

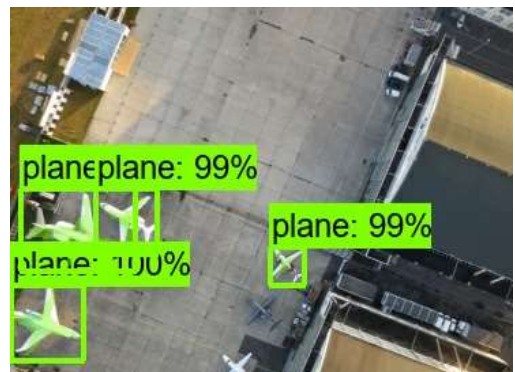

D

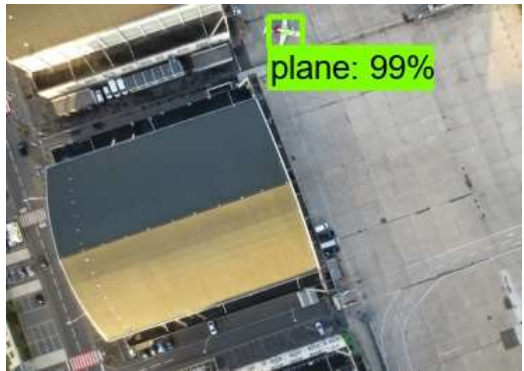
E

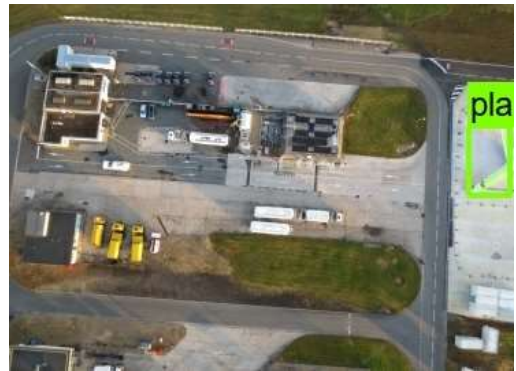
F

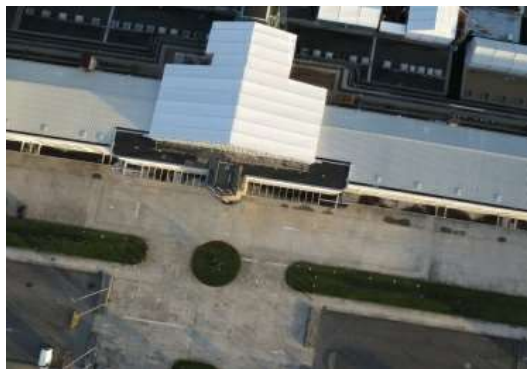
G

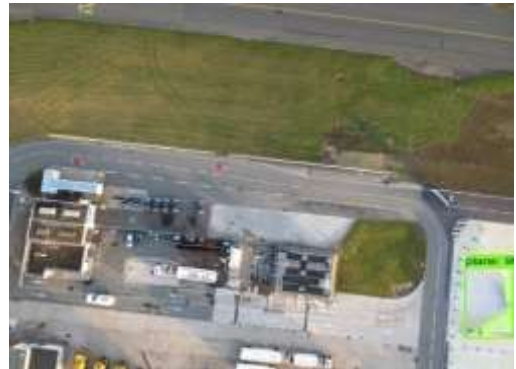
H

**Figure 11: Training Progress**

## REFERENCES

[1] E. Mazareanu. "Global air traffic - number of flights 2004-2021." Statista. (accessed 2020).

[2] L. Zhang, G. S. Xia, T. Wu, L. Lin, and X. C. Tai, "Deep Learning for Remote Sensing Image Understanding," *Journal of Sensors,* vol. 2016, 2016, doi: 10.1155/2016/7954154.

[3] X. Chen, S. Xiang, C. L. Liu, and C. H. Pan, "Aircraft detection by deep convolutional neural networks," *IPSJ Transactions on Computer Vision and Applications,* vol. 7, pp. 10-17, 2015, doi: 10.2197/ipsjtcva.7.10.

[4] C. Malladi, "Detection of Objects in Satellite images using Supervised and Unsupervised Learning Methods," 2017.

[5] A. Romero, C. Gatta, and G. Camps-Valls, "Unsupervised deep feature extraction for remote sensing image classification," *IEEE Transactions on Geoscience and Remote Sensing,* vol. 54, pp. 1349-1362, 2016, doi: 10.1109/TGRS.2015.2478379.

[6] J. Tang, C. Deng, G. B. Huang, and B. Zhao, "Compressed-domain ship detection on spaceborne optical image using deep neural network and extreme learning machine," *IEEE Transactions on Geoscience and Remote Sensing,* vol. 53, pp. 1174-1185, 2015, doi: 10.1109/TGRS.2014.2335751.

[7] X. Chen, S. Xiang, C. L. Liu, and C. H. Pan, "Vehicle detection in satellite images by hybrid deep convolutional neural networks," *IEEE Geoscience and Remote Sensing Letters,* vol. 11, pp. 1797-1801, 2014, doi: 10.1109/LGRS.2014.2309695.

[8] G. Cheng *et al.*, "Object detection in remote sensing imagery using a discriminatively trained mixture model," *ISPRS Journal of Photogrammetry and Remote Sensing,* vol. 85, pp. 32-43, 2013, doi: 10.1016/j.isprsjprs.2013.08.001.

[9] F. Bi, B. Zhu, L. Gao, and M. Bian, "A visual search inspired computational model for ship detection in optical satellite images," *IEEE Geoscience and Remote Sensing Letters,* vol. 9, pp. 749-753, 2012, doi: 10.1109/LGRS.2011.2180695.

[10] G. Hu, Z. Yang, J. Han, L. Huang, J. Gong, and N. Xiong, "Aircraft detection in remote sensing images based on saliency and convolution neural network," *Eurasip Journal on Wireless Communications and Networking,* vol. 2018, 2018, doi: 10.1186/s13638-018-1022-8.

[11] T. Tang, S. Zhou, Z. Deng, H. Zou, and L. Lei, "Vehicle detection in aerial images based on region convolutional neural networks and hard negative example mining," *Sensors (Switzerland),* vol. 17, 2017, doi: 10.3390/s17020336.

[12] S. Zhang, R. Wu, K. Xu, J. Wang, and W. Sun, "R-CNN-Based Ship Detection from High Resolution Remote Sensing Imagery," *Remote Sensing,* vol. 11, p. 631, 2019, doi: 10.3390/rs11060631.

[13] Z. Liu, L. Yuan, L. Weng, and Y. Yang, "A high resolution optical satellite image dataset for ship recognition and some new baselines," *ICPRAM 2017 - Proceedings of the 6th International Conference on Pattern Recognition Applications and Methods,* vol. 2017-Janua, pp. 324-331, 2017, doi: 10.5220/0006120603240331.

[14] U. Alganci, M. Soydas, and E. Sertel, "Comparative research on deep learning approaches for airplane detection from very high-resolution satellite images," *Remote Sensing,* vol. 12, 2020, doi: 10.3390/rs12030458.

[15] B. Traore, "Automatic airplane detection using deep learning techniques and very high-resolution satellite images," Master, Satellite Communication and Remote sensing, Istanbul Technical University, 2020.

[16] M. J. Khan, A. Yousaf, N. Javed, S. Nadeem, and K. Khurshid, "Automatic Target Detection in Satellite Images using Deep Learning," *Journal of Space Technology,* vol. 7, pp. 44-49, 2017.

[17] M. A. Nielsen, "Neural Networks and Deep Learning," ed: Determination Press, 2015.

[18] W. Li, S. Xiang, H. Wang, and C. Pan. Robust airplane detection in satellite images.

[19] H. Sun, X. Sun, H. Wang, Y. Li, and X. Li, "Automatic target detection in high-resolution remote sensing images using spatial sparse coding bag-of-words model," *IEEE Geoscience and*

Remote Sensing Letters, vol. 9, pp. 109-113, 2012, doi: 10.1109/LGRS.2011.2161569.

[20]  K. Cai, W. Shao, X. Yin, and G. Liu. Co-segmentation of aircrafts from high-resolution satellite images.

[21]  G. Liu, X. Sun, K. Fu, and H. Wang, "Aircraft recognition in high-resolution satellite images using coarse-to-fine shape prior," *IEEE Geoscience and Remote Sensing Letters,* vol. 10, pp. 573-577, 2013, doi: 10.1109/LGRS.2012.2214022.

[22]  J. W. Hsieh, J. M. Chen, C. H. Chuang, and K. C. Fan. Aircraft type recognition in satellite images.

[23]  J. Dai, K. He, and J. Sun, "Instance-Aware Semantic Segmentation via Multi-task Network Cascades," *Proceedings of the IEEE Computer Society Conference on Computer Vision and Pattern Recognition,* vol. 2016-Decem, pp. 3150-3158, 2016, doi: 10.1109/CVPR.2016.343.

[24]  R. Cao *et al.*, "Enhancing remote sensing image retrieval using a triplet deep metric learning network," *International Journal of Remote Sensing,* vol. 41, pp. 740-751, 2020, doi: 10.1080/2150704X.2019.1647368.

[25]  A. Filippidis, L. C. Jain, and N. Martin, "Fusion of intelligent agents for thé detection of aircraft in SAR images," *IEEE Transactions on Pattern Analysis and Machine Intelligence,* vol. 22, pp. 378-384, 2000, doi: 10.1109/34.845380.

[26]  D. Marmanis, M. Datcu, T. Esch, and U. Stilla, "Deep learning earth observation classification using ImageNet pretrained networks," *IEEE Geoscience and Remote Sensing Letters,* vol. 13, pp. 105-109, 2016, doi: 10.1109/LGRS.2015.2499239.

[27]  M. Castelluccio, G. Poggi, C. Sansone, and L. Verdoliva, "Land Use Classification in Remote Sensing Images by Convolutional Neural Networks."

[28]  A. F.-S. L. A. Ruiz and J. A. Recio, "TEXTURE FEATURE EXTRACTION FOR CLASSIFICATION OF REMOTE SENSING DATA USING WAVELET DECOMPOSITION: A COMPARATIVE STUDY," ed.

[29]  G. Camps-Valls and A. Rodrigo-González, "Classification of satellite images with regularized AdaBoosting of RBF neural networks," *Studies in Computational Intelligence,* vol. 83, pp. 307-326, 2008, doi: 10.1007/978-3-540-75398-8_14.

[30]  G. Sheng, W. Yang, T. Xu, and H. Sun, "High-resolution satellite scene classification using a sparse coding based multiple feature combination," *International Journal of Remote Sensing,* vol. 33, pp. 2395-2412, 2012, doi: 10.1080/01431161.2011.608740.

[31]  G. Cheng and J. Han, "A survey on object detection in optical remote sensing images," *ISPRS Journal of Photogrammetry and Remote Sensing,* vol. 117, pp. 11-28,, doi: 10.1016/j.isprsjprs.2016.03.014.

[32]  S. Chen, R. Zhan, and J. Zhang, "Geospatial Object Detection in Remote Sensing Imagery Based on Multiscale Single-Shot Detector with Activated Semantics," *Remote Sensing,* vol. 10, p. 820, 2018, doi: 10.3390/rs10060820.

[33]  X. Yang *et al.*, "Automatic Ship Detection in Remote Sensing Images from Google Earth of Complex Scenes Based on Multiscale Rotation Dense Feature Pyramid Networks," *Remote Sensing,* vol. 10, p. 132, 2018, doi: 10.3390/rs10010132.

[34]  T. N. Mundhenk, G. Konjevod, W. A. Sakla, and K. Boakye. A large contextual dataset for classification, detection and counting of cars with deep learning.

[35]  P. Zhou, G. Cheng, Z. Liu, S. Bu, and X. Hu, "Weakly supervised target detection in remote sensing images based on transferred deep features and negative bootstrapping," *Multidimensional Systems and Signal Processing,* vol. 27, pp. 925-944, 2016, doi: 10.1007/s11045-015-0370-3.

[36]  M. W. Gardner and S. R. Dorling, "Artificial neural networks (the multilayer perceptron) - a review of applications in the atmospheric sciences," *Atmospheric Environment,* vol. 32, pp. 2627-2636, 1998, doi: 10.1016/S1352-2310(97)00447-0.

[37]  W. Liu *et al.* SSD: Single shot multibox detector.

[38]  W. Li, G. Wu, F. Zhang, and Q. Du, "Hyperspectral Image Classification Using Deep Pixel-Pair Features," *IEEE Transactions on Geoscience and Remote Sensing,* vol. 55, pp. 844-853, 2017, doi: 10.1109/TGRS.2016.2616355.

[39]  L. Mou, P. Ghamisi, and X. X. Zhu. Fully conv-deconv network for unsupervised spectral-spatial feature extraction of hyperspectral imagery via residual learning.

[40]  L. Mou, P. Ghamisi, and X. X. Zhu, "Unsupervised spectral-spatial feature learning via deep residual conv-deconv network for hyperspectral image classification," *IEEE Transactions on Geoscience and Remote Sensing,* vol. 56, pp. 391-406, 2018, doi: 10.1109/TGRS.2017.2748160.

[41]  J. Fan, T. Chen, and S. Lu, "Unsupervised Feature Learning for Land-Use Scene Recognition," *IEEE Transactions on Geoscience and Remote Sensing,* vol. 55, pp. 2250-2261, 2017, doi: 10.1109/TGRS.2016.2640186.

[42]  D. Dai and W. Yang, "Satellite image classification via two-layer sparse coding with biased image representation," *IEEE Geoscience and Remote Sensing Letters,* vol. 8, pp. 173-176, 2011, doi: 10.1109/LGRS.2010.2055033.

[43]  L. A. Ruiz, A. Fdez-Sarría, and J. A. Recio, "TEXTURE FEATURE EXTRACTION FOR CLASSIFICATION OF REMOTE SENSING DATA USING WAVELET DECOMPOSITION: A COMPARATIVE STUDY."

[44]  C. S. Li and V. Castelli. Deriving texture feature set for content-based retrieval of satellite image database.

[45]  L. Bruzzone and L. Carlin, "A multilevel context-based system for classification of very high spatial resolution images," *IEEE Transactions on Geoscience and Remote Sensing,* vol. 44, pp. 2587-2600, 2006, doi: 10.1109/TGRS.2006.875360.

[46]  S. A. Wagner, "SAR ATR by a combination of convolutional neural network and support vector machines," *IEEE Transactions on Aerospace and Electronic Systems,* vol. 52, pp. 2861-2872, 2016, doi: 10.1109/TAES.2016.160061.

[47]  Q. Zhao and J. C. Principe, "Support vector machines for SAR automatic target recognition," *IEEE Transactions on Aerospace and Electronic Systems,* vol. 37, pp. 643-654, 2001, doi: 10.1109/7.937475.

[48]  J. Deng, W. Dong, R. Socher, L.-J. Li, K. Li, and L. Fei-Fei. ImageNet: A large-scale hierarchical image database.

[49]  H. C. Shin *et al.*, "Deep Convolutional Neural Networks for Computer-Aided Detection: CNN Architectures, Dataset Characteristics and Transfer Learning," *IEEE Transactions on Medical Imaging,* vol. 35, pp. 1285-1298, 2016, doi: 10.1109/TMI.2016.2528162.

[50]  D. I. Moldovan and C.-I. Wu, "A hierarchical knowledge based system for airplane classification," *IEEE transactions on software engineering,* vol. 14, no. 12, pp. 1829-1834, 1988.

[51]  M. Pesaresi and J. A. Benediktsson, "A new approach for the morphological segmentation of high-resolution satellite imagery," *IEEE transactions on Geoscience and Remote Sensing,* vol. 39, no. 2, pp. 309-320, 2001.

[52]  S. Greenberg and H. Guterman, "Neural-network classifiers for automatic real-world aerial image recognition," *Applied optics,* vol. 35, no. 23, pp. 4598-4609, 1996.

[53]  M. Abadi *et al.*, "Tensorflow: A system for large-scale machine learning," in *12th {USENIX} symposium on operating systems design and implementation ({OSDI} 16)*, 2016, pp. 265-283.

[54]  J. Redmon and A. Farhadi, "Yolov3: An incremental improvement," *arXiv preprint arXiv:1804.02767,* 2018.

[55]  C. Wojek, G. Dorkó, A. Schulz, and B. Schiele, "Sliding-windows for rapid object class localization: A parallel technique," in *Joint Pattern Recognition Symposium*, 2008: Springer, pp. 71-81.

[56]  G.-H. Nie, P. Zhang, X. Niu, Y. Dou, and F. Xia, "Ship detection using transfer learned single shot multi box detector," in *ITM Web of Conferences*, 2017, vol. 12: EDP Sciences, p. 01006.

[57]  M. Radovic, O. Adarkwa, and Q. Wang, "Object recognition in aerial images using convolutional neural networks," *Journal of Imaging,* vol. 3, no. 2, p. 21, 2017.

[58]  N. Ammour, H. Alhichri, Y. Bazi, B. Benjdira, N. Alajlan, and M. Zuair, "Deep learning approach for car detection in UAV imagery," *Remote Sensing,* vol. 9, no. 4, p. 312, 2017.

[59]  M. Li, Z. Zhang, L. Lei, X. Wang, and X. Guo, "Agricultural Greenhouses Detection in High-Resolution Satellite Images Based on Convolutional Neural Networks: Comparison of Faster

R-CNN, YOLO v3 and SSD," *Sensors,* vol. 20, no. 17, p. 4938, 2020.

[60] K. Reda and M. Kedzierski, "Detection, Classification and Boundary Regularization of Buildings in Satellite Imagery Using Faster Edge Region Convolutional Neural Networks," *Remote Sensing,* vol. 12, no. 14, p. 2240, 2020.

[61] S.-J. Hong, Y. Han, S.-Y. Kim, A.-Y. Lee, and G. Kim, "Application of deep-learning methods to bird detection using unmanned aerial vehicle imagery," *Sensors,* vol. 19, no. 7, p. 1651, 2019.

[62] K. Nguyen, N. T. Huynh, P. C. Nguyen, K.-D. Nguyen, N. D. Vo, and T. V. Nguyen, "Detecting Objects from Space: An Evaluation of Deep-Learning Modern Approaches," *Electronics,* vol. 9, no. 4, p. 583, 2020.

[63] S. Zhang, Y. Wu, C. Men, and X. Li, "Tiny YOLO optimization oriented bus passenger object detection," *Chinese Journal of Electronics,* vol. 29, no. 1, pp. 132-138, 2020.

[64] sensefly. *Dataset of Le Bourget airport in Paris*, Parrot Group, 2017,

[65] Tzutalin. "LabelImg a graphical image annotation tool." github. (accessed 2020).

[66] Tzutalin. "LabelImg a graphical image annotation tool." github. (accessed 2020).

[67] P. D. K. He G. Gkioxari and R. Girshick. Mask r-cnn.

[68] G. Cheng and J. Han, "A survey on object detection in optical remote sensing images," in *ISPRS Journal of Photogrammetry and Remote Sensing* vol. 117, ed: Elsevier B.V., 2016, pp. 11-28.

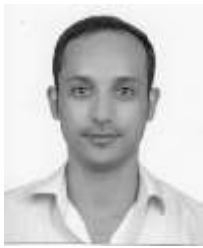

**W.T. Alshaibani** received the B.S. degree in electrical engineering from Yarmouk University, Irbid, Jordan, in 2015 and M.Cs in satellite communication and remote sensing from Istanbul technical university (ITU), Istanbul, Turkey. He is currently pursuing the PhD degree in Electrical engineering at ITU. His research interests include 5G and deep learning fields.

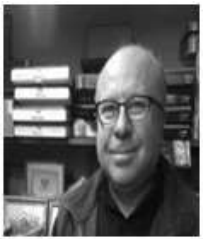

**Mustafa Helvaci** received the B.S., M.S. degree and PhD degrees from Ankara University in 1989, 1994 and 2003 consequently. He works on electromagnetic waves, RF, MW and IR theory, telecommunication, radiation material interactions, radiation and heat transfer, molecular spectroscopy and chemical abundance analysis with remote sensing techniques
.

**IBRAHEEM SHAYEA** received the B.Sc. degree in electronic engineering from the University of Diyala, Baqubah, Iraq, in 2004, and the M.Sc. degree in computer and communication engineering and the Ph.D. degree in mobile communication engineering from The National University of Malaysia, Universiti Kebangsaan Malaysia (UKM), Malaysia, in 2010 and 2015, respectively. Since the 1st of January 2011 until 28 February 2014, he has been Research and a Teaching Assistant with Universiti Kebangsaan Malaysia (UKM), Malaysia. Then, from the 1st of January 2016 until 30 June 2018, he joined Wireless Communication Center (WCC), University of Technology Malaysia (UTM), Malaysia, and worked there as a Research Fellow. He is currently a Researcher Fellow with Istanbul Technical University (ITU), Istanbul, Turkey, since the 1st of September 2018 until now. His main research interests include in wireless communication systems, mobility management, radio propagation, and the Internet of Things (IoT).

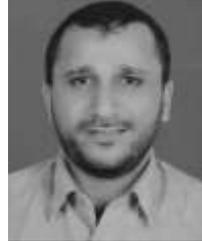

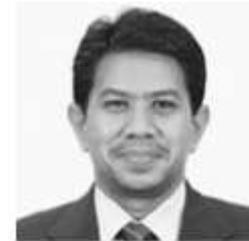

**Hafizal Mohamad**(SM'10) received the B.Eng. degree (honors) and the Ph.D. degree in electronic engineering from the University of Southampton, UK in 1998 and 2003, respectively. He is currently a Professor at the Faculty of Engineering and Built Environment, Universiti Sains Islam Malaysia. His research interests include wireless communication, cognitive radio, mesh networks and Internet of Things. He is the co-inventor of 36 filed patents and 9 granted patents in the field of wireless communication. He has coauthored over 100 research papers. He has served in various leadership roles in the IEEE, including the Vice Chair for IEEE Malaysia Section (2013) and the Chair for IEEE ComSoc/VT Chapter (2009-11). He was the Conference Operation Chair for the IEEE ICC 2016, the Technical Program Chair for the IEEE VTC Spring 2019 and APCC 2011. He is the recipient of several awards, including the ASEAN Outstanding Scientist and Technologist Award (AOSTA) and Top Research Scientist Malaysia (TRSM) by Academy of Sciences Malaysia. He is a registered Professional Engineer with the Board of Engineers Malaysia. He is an enthusiastic supporter of industrial and academic liaison. He is also appointed as an expert panel for various technical committees.